# Smart Attendance System Usign CNN

Shailesh Arya
Dept. of Computer Science
Pandit Deendayal Petoleum University
Gandhinagar,India
aryashailesh@outlook.com

Hrithik Mesariya
Dept. of Computer Science
Pandit Deendayal Petroleum University
Gandhinagar,India
hrithik.mce16@sot.pdpu.ac.in

Vishal Parekh
Dept. of Computer Science
Pandit Deendayal Petroleum University
Gandhinagar,India.
vishal.pce16@sot.pdpu.ac.in

***Abstract*—The research on the attendance system has been going for a very long time, numerous arrangements have been proposed in last decade to make this system efficient and less time consuming, but all those systems have several flaws. In this paper we are introducing a smart and efficient system for attendance using face detection and face recognition. This system can be used to take attendance in colleges or offices using real time face recognition with the help of Convolution Neural Network(CNN). The conventional methods like Eigen faces and Fisher faces are sensitive to lighting, noise, posture, obstruction, illumination etc. Hence, we have used CNN to recognize the face and overcome such difficulties. The attendance records will be updated automatically and stored in excel sheet as well as in database. We have used the MongoDB as backend database for attendance records.**

## I. INTRODUCTION

Attendance is the measure of an individual's presence at his work place. Attendance plays an important role in our regular lives. Traditional way of taking attendance in schools/colleges or even in business firm is by manually marking attendance in registers. This is complete waste of time and effort for both students and lecturers as well. With advancement of technology, some other ways introduced to take attendance. One popular method for attendance is RFID based attendance system [1], which uses electromagnetic fields to automatically identify and track tags attached to person. Despite being fast and robust system, it violates the privacy and security of human beings. Now a day, attendance using biometric i.e. finger print recognition [2], Iris scanning [3] etc. are used. Finger print recognition based attendance system is one of the most successful system available and is used by mostly all the business firms these days. Iris scanning based attendance is more secure but it takes greater time. All the mentioned systems for attendance gets advanced with time but still they can be fooled by someone and also required some sort of human intervention to manage system. In order to overcome such shortcomings of traditional attendance, we proposed a more convenient and robust system called Smart Attendance System based on face recognition [7] is one of the most secure and advanced system. The proposed system uses the special type of Convolution Neural Network(CNN) known as Siamese Network. CNN is basically a Deep Neural Network (i.e. Neural Network having more layers) [5] in which lower layers contain sparse features which are appropriate for learning landmark localization. On the other hand, deeper layers contain dense features for the learning of complex task such as face detection and gender recognition.

The face recognition process is divided into two parts: Face Detection and Face Recognition using Datasets. Face detection is detecting the presence of the face and face location in the image or live video frame. We have used HAAR Cascade classifier proposed by Viola and Jones to detect faces. The Haar classifier [6] is a machine learning based approach which is trained from many positive images (i.e. images with faces) and negative images (i.e. images without faces). It uses Haar features to extract features from image.

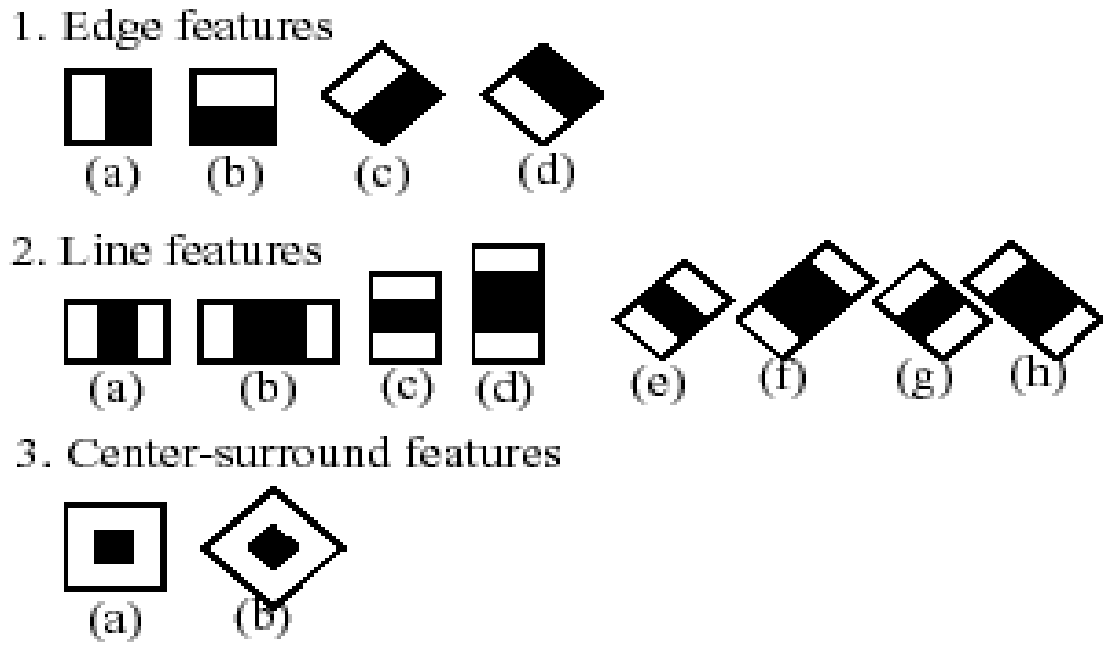

Fig-1. Haar-like Features

Fig-1. Shows some of the features of Haar Classifier. Each window is placed on the image to calculate a single feature by subtracting the sum of pixels under white part of window from the pixels under the black part of window. In the end algorithm considers the most important features (i.e. eyes, nose, lips) to detect face.

For Face Recognition, we have trained a special CNN model known as Siamese Network [4]. The word "Siamese" means joint or connected. The Siamese network is the neural network architecture that contain two or more identical (i.e. same configuration with same parameters and weights) subnetworks. It learns what make 2 pair of inputs the same (i.e. dog-dog, eel-eel).

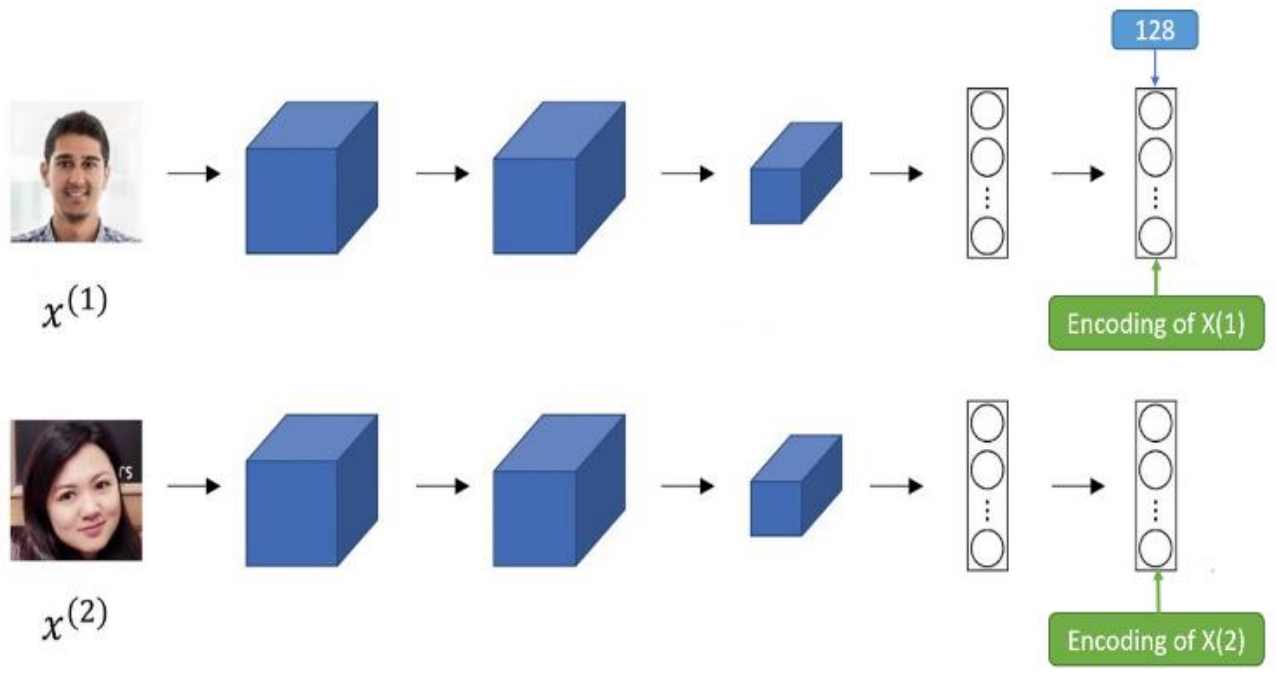

Fig-2. Siamese Network general Architecture

Since we are going to recognize the faces, we will utilize a Convolutional Siamese Network. Fig.-2 shows the general architecture of Siamese Network. It consists of convolution layers, max-pooling layers and fully connected layer. The algorithm will work in following manner:

1) We take two images ($X^{(1)}$, $X^{(2)}$). Both are feed to the Convolution Neural Network (CNN).

2) The last layer of the CNN will produce a fixed size vector (embeddings of the image). Since we have fed

two images, we will get two embeddings ($f(X^{(1)})$, $f(X^{(2)})$).

3) The absolute distance between the vectors is calculated.

4) If the distance is very small then we classify that the two images are same otherwise they are different.

The loss can be calculated using different loss function. Some popular loss functions are binary cross-entropy, triplet loss, absolute distance etc. In this implementation we have used absolute distance loss function to train the model through back propagation.

## II. PROPOSED METHODOLOGY

### A. *Block Digram*

Fig.-3 depicts the functional block of our Smart Attendance System. The system uses live camera feed from webcam to detect faces in real-time and do the necessary preprocessing to compute the Face encoding using the pretrained Siamese Network.

For training the model, we take 50 different images of person. Initially Face Detection using Haar classifier is applied to detect faces and stored in database which is followed by the Siamese network to compute the unique 128-bit encodings.

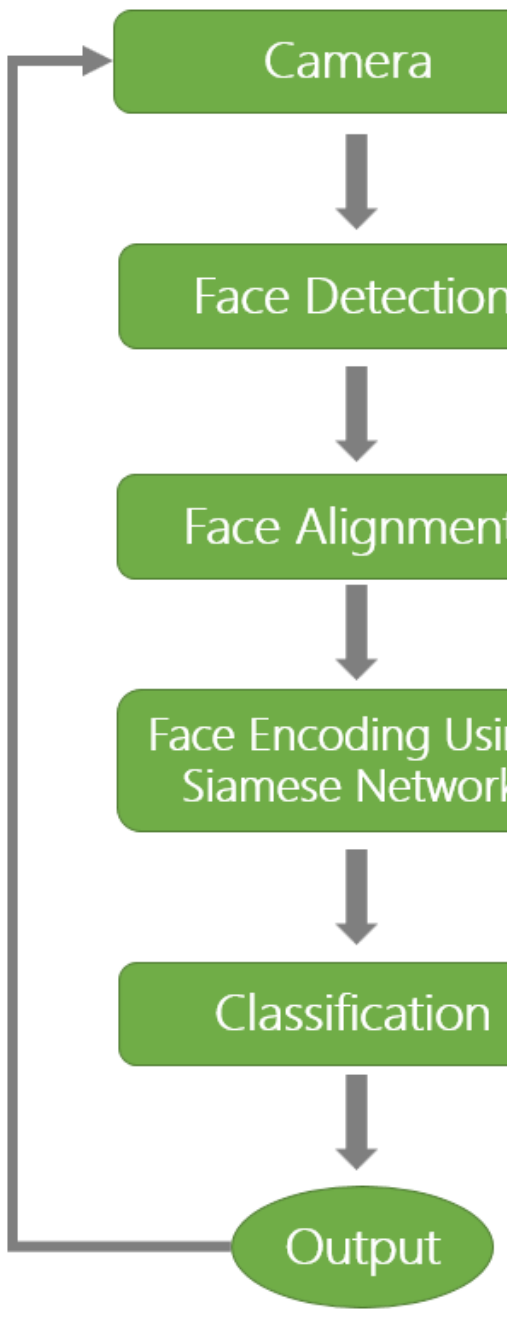

Fig-3. Data Flow Diagram

Once the 128-bit encodings computed then the output in sent to another neural network which outputs the probability of the face to be one of the persons present in dataset.

### B. *Methodology*

For implementing the Smart Attendance System using face recognition, we have followed the following steps in same order:

1) Enrollment of Students
    a. Face Detection
    b. Preprocessing
    c. Database Creation
2) Train the model
3) Marking attendance
4) Store attendance to the database

**1) Enrollment of Person:**

The person will be enrolled to the database with their name and unique id number. We take 50 different images of person using live webcam feed. Images will be stored in a folder with the person's id as folder name.

**a. Face Detection:** We have used Haar cascade classifier available in OpenCV to detect face in real time.

**b. Preprocessing:** Preprocessing includes cropping and aligning the face of person to the center of the frame.

**c. Database Creation:** Different 50 images of same person is stored in a folder with person's id number as folder name.

**2) Train the Model:**

This network is pretrained on a pretty large dataset and produces very accurate and almost unique 128-dimensional vector for a particular face given that the images fed to it are cropped and faces are aligned. The network expects the input size to be 160 X 160 X 3. The model can work with Colored images as well.

**3) Marking Attendance:**

The second neural network is used for classification task. It has dense architecture which takes the 128-dimensional vector as input and outputs the probability of the face to one of the persons.

**4) Store Attendance to Database:**

We have used the MongoDB as backend database. Once the attendance is taken by identifying the face in real time, a counter is incremented in database. Also, after the end of attendance task, it generates a csv filed which houses the attendance records of Students with their names and unique id numbers.

## III. IMPLEMENTATION

The proposed method is implemented on Python programming language with complementary tools like eel to run python script from webpage. Eel is a little Python library for making simple Electron-like HTML/JS GUI Apps. It hosts a local webserver, then provides features to communicate between JavaScript and Python. MongoDB is used to store the attendance record.

- Application GUI

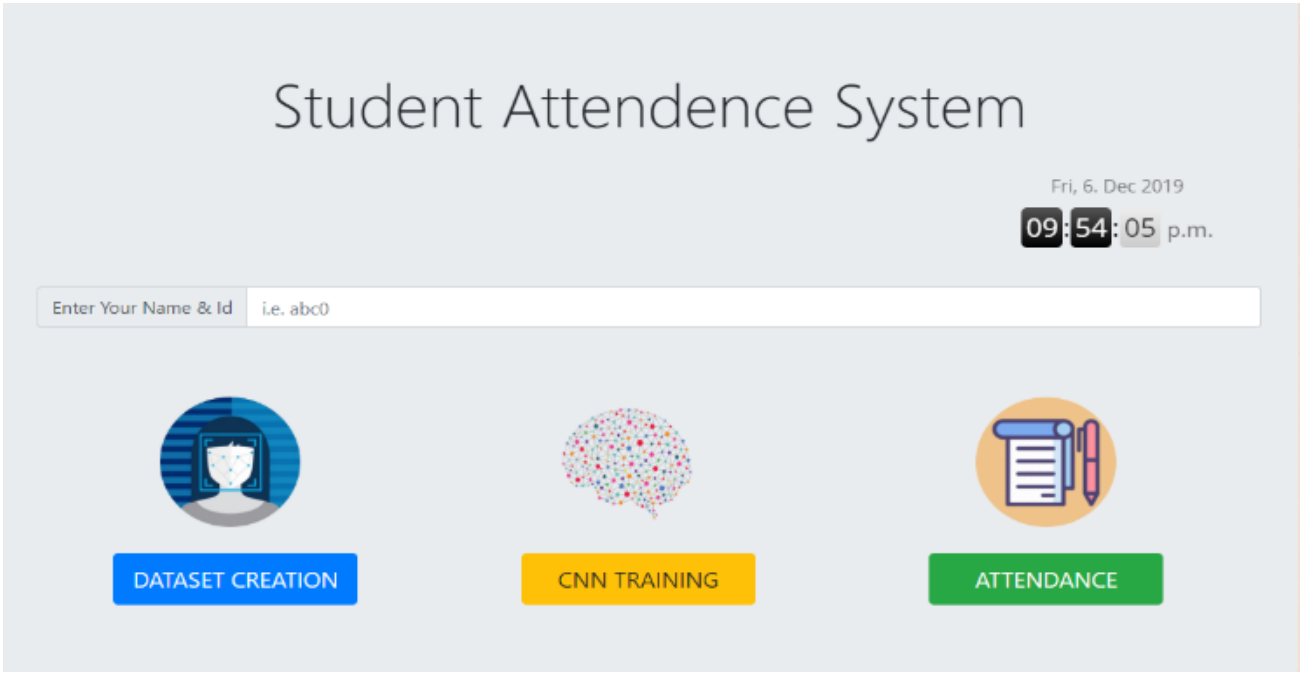

Fig-4. Application GUI using Bootstrap

Fig-4. Shows the Application GUI which is developed with help of HTML and Bootstrap. This will be used to interact with the system. It has basically three HTML buttons to leverage the features of Smart Attendance System. The first button is for creating the Dataset of particular person. Person has to insert his name and id (i.e. abc0) and then the system will capture his/her photos with help of high-definition webcam. A folder with user's name and id will be created and stored in system for further training of model. Fig-5 a. Shows the dataset folders and Fig-5 b. inside view of dataset.

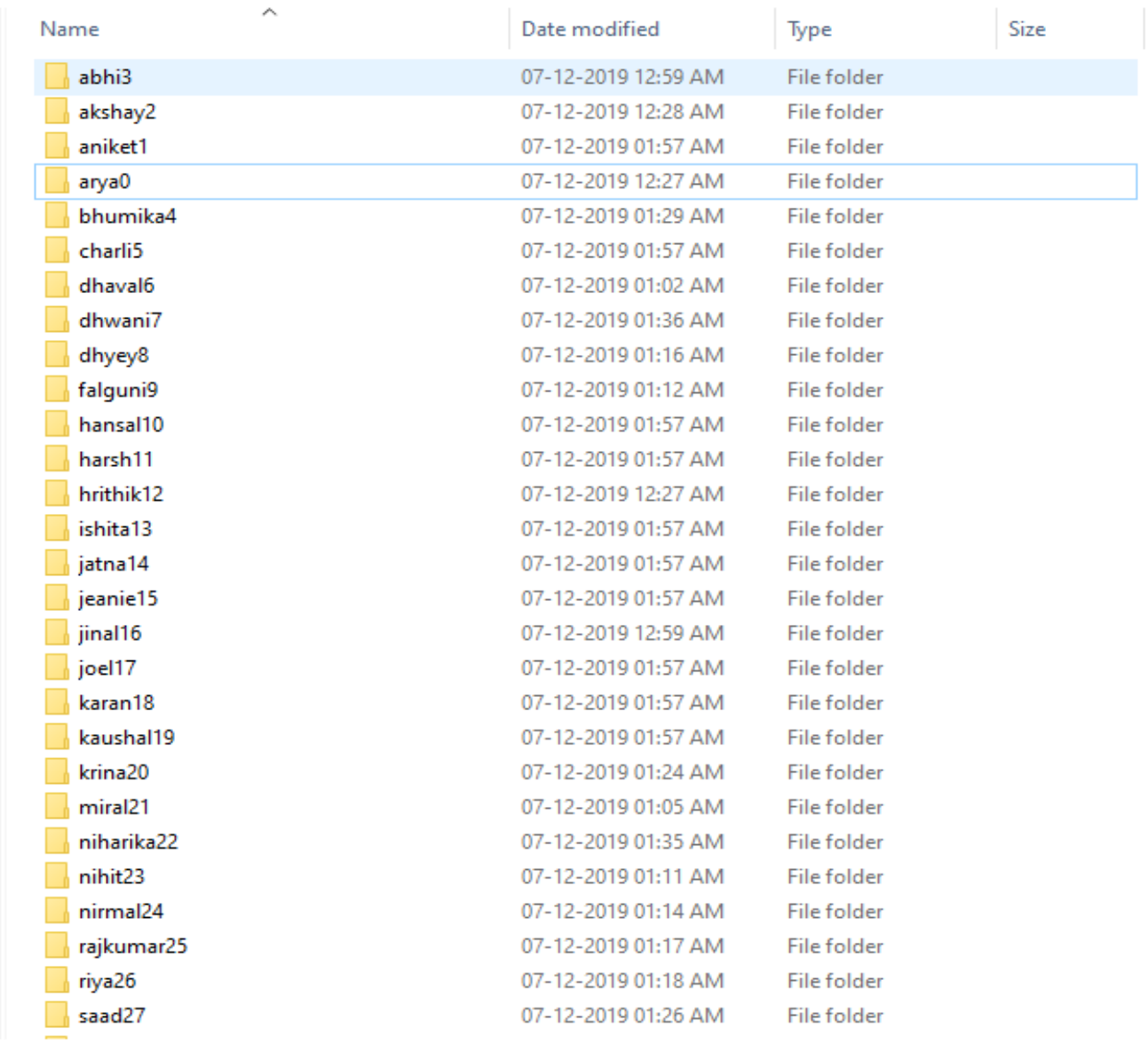

Fig-5 a. Dataset overview

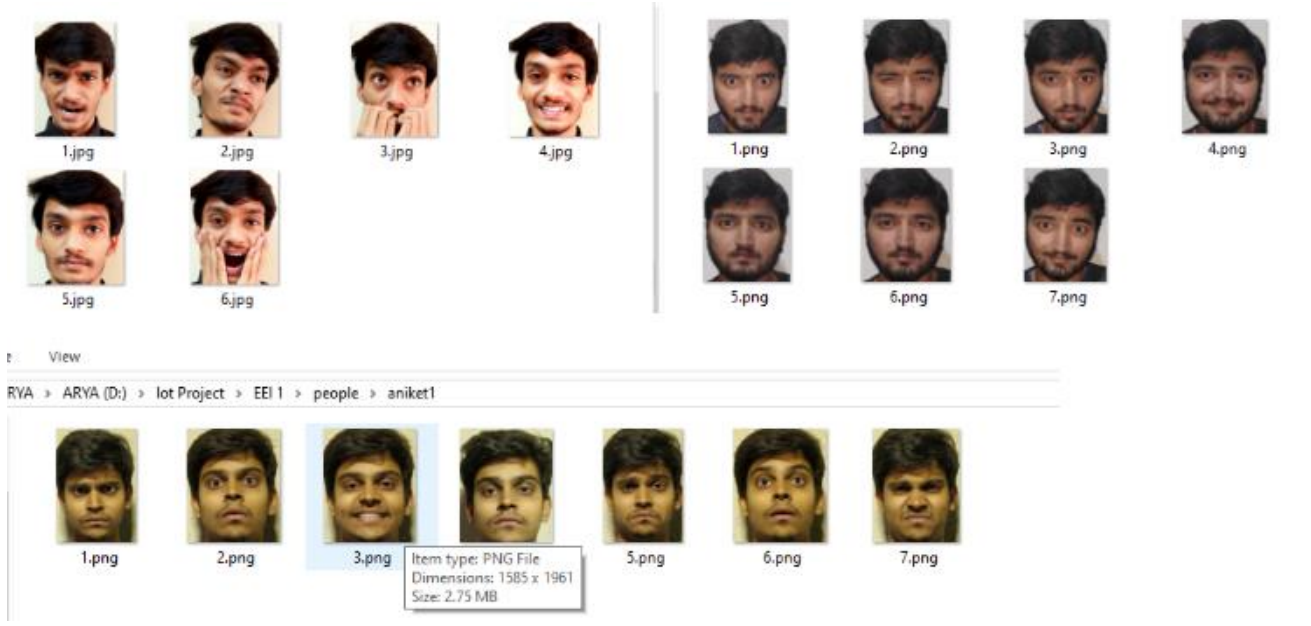

Fig-5 b. Dataset of different people

The Second button is to train the model. Initially it will look in the dataset directory and for each images of different person, it will compute the 128 dimensional vector which is then divided into training and testing dataset to train the model. The dataset creation and training model is one-time process which we have to perform. After training and testing, we have got pretty good accuracy of 94 percentages.

The last button is for taking attendance. Below Fig-6 - a and Fig.6 – b shows the attendance taking in real time.

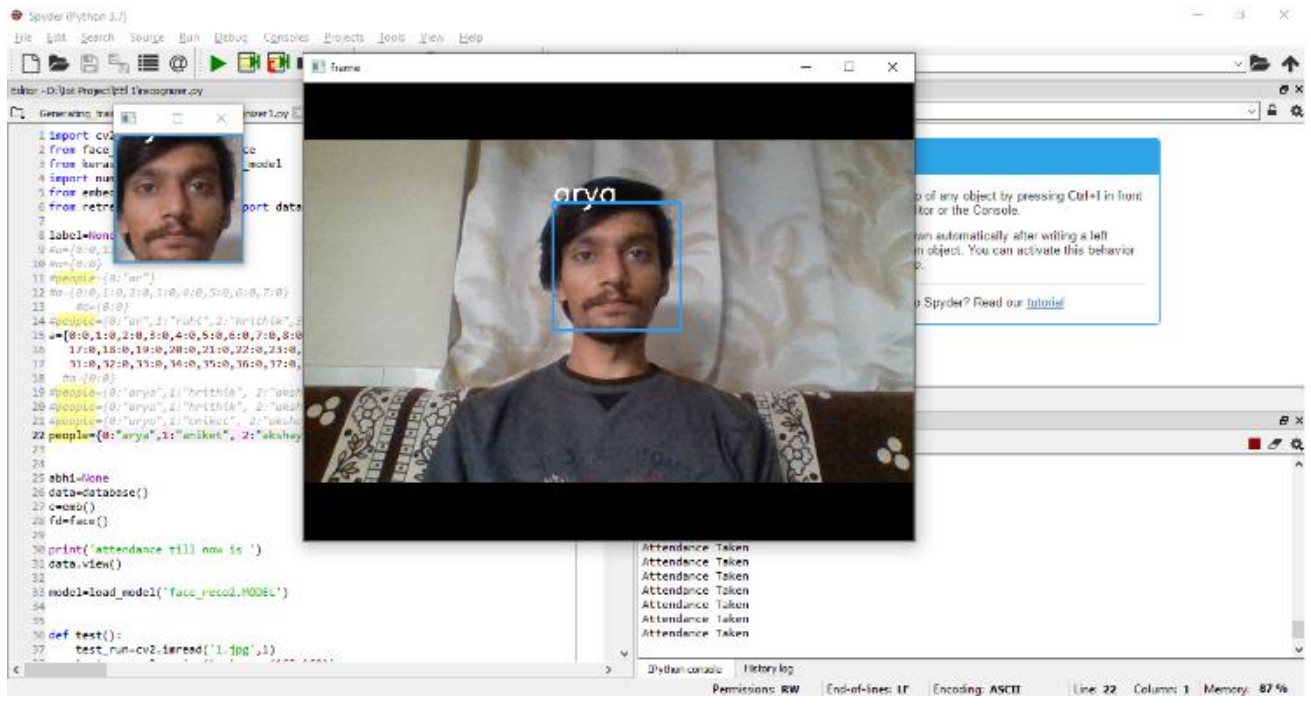

Fig-6-a. Taking Attendance in Real Time

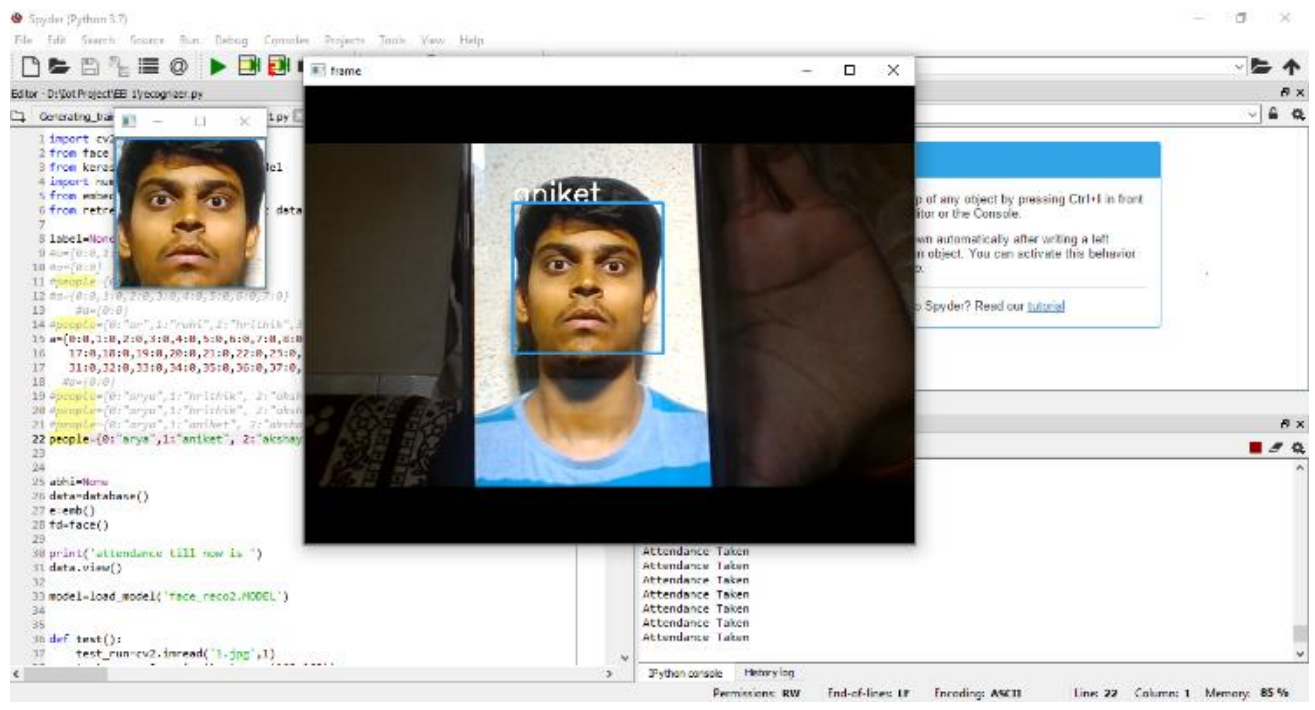

Fig-6-b. Taking Attendance in Real Time

This will launch the camera module, which will detect the face of person in real time and compute face encoding of the detected face and the it will compare the encodings with the encodings which we have computed during training the model. Basically, it will find the distance between the encodings and it will consider the person's encoding corresponds to the least distance.

Finally, the counter will increment and updated in MongoDB database. Fig.7 shows the stored attendance results in MongoDB.

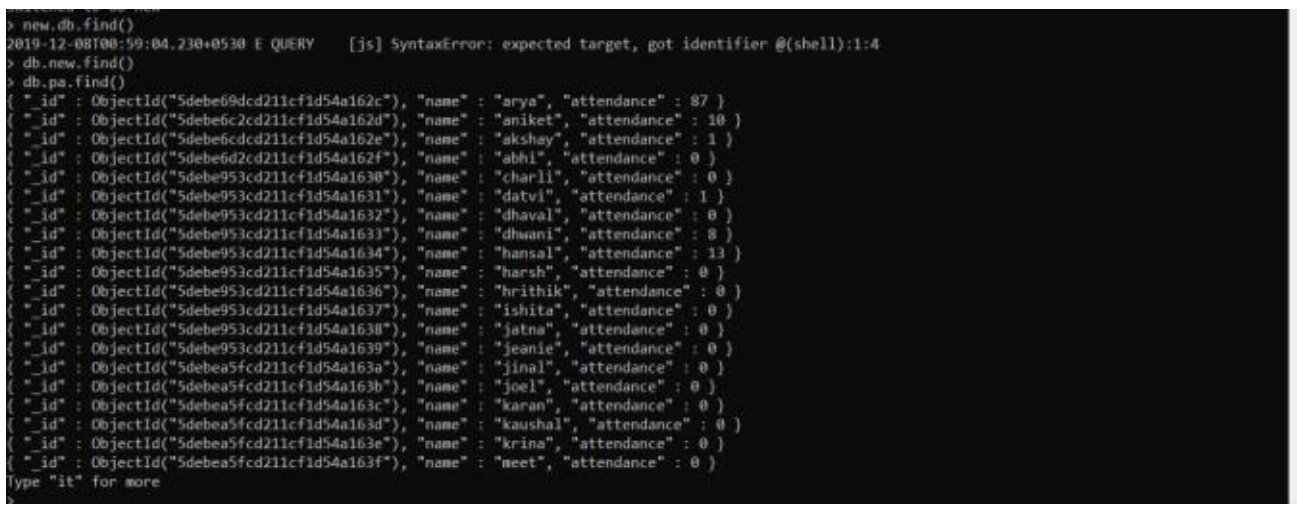

Fig-7. Stored Attendance Database in MongoDB

- TESTED ENVIRONMENT

| NO. | Resource Type | Details |
|---|---|---|
| 1 | Host O.S. | Windows 10, 64 bits |
| 2 | CPU | Intel(R) Core (TM) i5-2100 CPU @ 3.10GHz |
| 3 | RAM | 8.0 GB |
| 4 | HDD | 1 TB |
| 5 | Type of HDD | HDD |
| 6 | Webcam Resolution | 1280 X 720 |
| 7 | Keras version | 2.3.1 |
| 8 | Tensor flow Version | 1.14.0 |